\RequirePackage{fix-cm}
\documentclass[twocolumn]{svjour3}          %
\smartqed  %
\usepackage{graphicx}
\usepackage{amsmath}
\usepackage{hyperref}
\usepackage{url}
\usepackage{natbib}
\usepackage{amsmath,amssymb,amsfonts}
\usepackage[ruled,linesnumbered]{algorithm2e}
\usepackage{algorithmic}
\usepackage{booktabs,multirow,multicol}
\usepackage{xcolor}

\usepackage{caption}
\begin{document}
\sloppy

\title{
3D CoCa v2: Contrastive Learners with Test-Time Search for Generalizable Spatial Intelligence}

\author{Hao Tang$^{*\dag}$ \and
        Ting Huang$^*$ \and
        Zeyu Zhang$^*$
}

\authorrunning{Hao Tang, Ting Huang, and Zeyu Zhang} %

\institute{Hao Tang \at
              School of Computer Science, Peking University \\
              \email{bjdxtanghao@gmail.com}           %
           \and
           Ting Huang \at
              School of Computer Science, Peking University \\
              \email{hting247@gmail.com}
           \and
           Zeyu Zhang \at
              School of Computer Science, Peking University \\
              \email{steve.zeyu.zhang@outlook.com} \\
              ~\\
           $^*$Equal contribution.  $\dag$Corresponding author.
}

\date{Received: date / Accepted: date}

\maketitle

\begin{abstract}
Spatial intelligence refers to the ability to perceive, reason about, and describe objects and their relationships within three-dimensional environments, forming a foundation for embodied perception and scene understanding.
3D captioning aims to describe 3D scenes in natural language; however, it remains challenging due to the sparsity and irregularity of point clouds and, more critically, the weak grounding and limited out-of-distribution (OOD) generalization of existing captioners across drastically different environments, including indoor and outdoor 3D scenes.
To address this challenge, we propose \textbf{3D CoCa v2}, a generalizable 3D captioning framework that unifies contrastive vision-language learning with 3D caption generation and further improves robustness via test-time search (TTS) without updating the captioner parameters.
3D CoCa v2 builds on a frozen CLIP-based semantic prior, a spatially-aware 3D scene encoder for geometry, and a multimodal decoder jointly optimized with contrastive and captioning objectives, avoiding external detectors or handcrafted proposals. At inference, TTS produces diverse caption candidates and performs reward-guided selection using a compact scene summary.
Experiments show improvements over 3D CoCa of \textbf{+1.50} CIDEr@0.5IoU on ScanRefer and \textbf{+1.61} CIDEr@0.5IoU on Nr3D, and \textbf{+3.8} CIDEr@0.25 in zero-shot OOD evaluation on TOD$^{3}$Cap.
Code will be released at \url{https://github.com/AIGeeksGroup/3DCoCav2}.
\end{abstract}

\begin{figure}[ht]
    \centering
    \includegraphics[width=\linewidth]{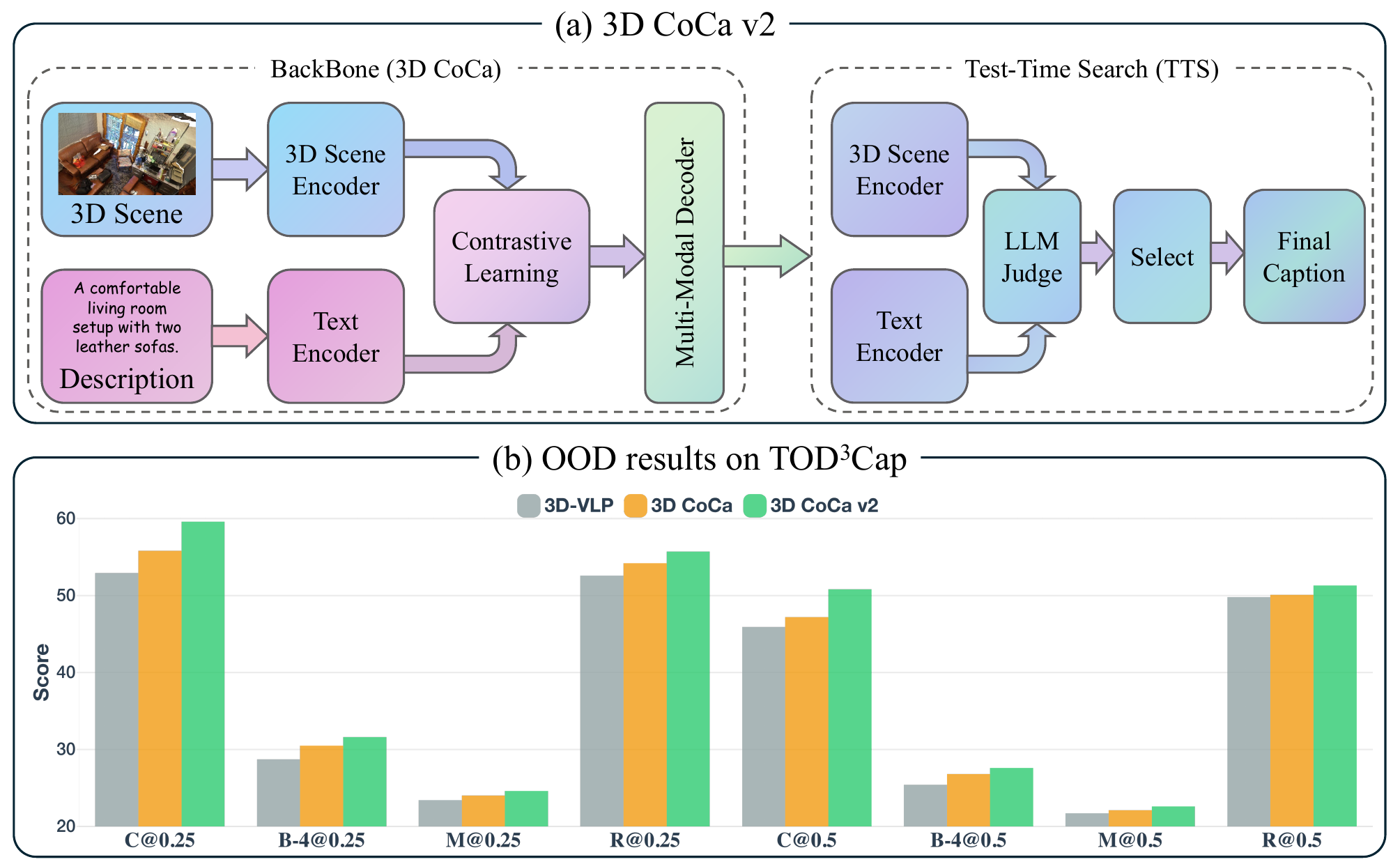}
    \caption{\textbf{Overview of 3D CoCa v2 and OOD results on TOD$^3$Cap.}
    (a) 3D CoCa v2 extends 3D CoCa~\citep{huang20253dcoca} with an inference-only test-time search (TTS) module and an external LLM judge.
    (b) Zero-shot OOD performance on TOD$^3$Cap~\citep{todocap2025} comparing 3D-VLP~\citep{3dvlp2024}, 3D CoCa~\citep{huang20253dcoca}, and 3D CoCa v2 under standard captioning metrics at IoU 0.25 and 0.5.
    }
    \label{fig:main}
    \vspace{-0.4cm}
\end{figure}

\section{Introduction}
\label{sec:intro}
Developing spatial intelligence in real-world environments requires models that can not only perceive 3D geometry but also communicate spatial understanding through natural language.
In recent years, 3D representation learning has attracted increasing attention due to its broad impact on robotics, autonomous driving, and augmented reality~\citep{sportscap2021,chen2021tightcap}. In parallel, the convergence of computer vision and natural language processing has fostered vision-language tasks that connect perception with linguistic understanding, where captioning serves as an intuitive interface for interpreting complex scenes. Although large-scale vision-language models have led to substantial progress in 2D captioning, extending captioning to 3D remains considerably more challenging: point clouds are sparse and irregular, objects are cluttered or partially observed, and faithful descriptions require not only recognizing object attributes but also reasoning about their spatial context.
Early 3D captioning methods, therefore, largely adopted a two-stage ``detect-then-describe'' paradigm, first generating object proposals and then describing each region. Scan2Cap~\citep{scan2cap_2021} is an early representative that cascades 3D detection and caption generation, followed by efforts that incorporate language pre-training and cross-modal alignment to improve 3D captioning quality~\citep{camm2023}.

Despite their effectiveness, two-stage pipelines have well-known drawbacks: the detection stage often produces redundant or noisy proposals and requires additional post-processing, such as Non-Maximum Suppression~\citep{nms2006}. Meanwhile, the quality of captions becomes tightly coupled to detection accuracy.
To alleviate these issues, one-stage end-to-end frameworks have gained popularity.
Vote2Cap-DETR~\citep{vote2cap2023} and Vote2Cap-DETR++ \citep{vote2cap++2024} adopt Transformer-based formulations that jointly localize and describe objects. Recent designs such as BiCA \citep{bica2025} and See-It-All~\citep{seeitall2024} further enhance contextual aggregation in 3D scenes.
Meanwhile, TOD$^3$Cap~\citep{todocap2025} targets outdoor settings and highlights the growing need to handle diverse real-world environments.
However, existing 3D captioners still face two fundamental challenges, especially under OOD deployment: (i) their grounding degrades markedly when moving beyond the training domain. For example, models trained on indoor RGB-D reconstructions often encounter drastically different geometries, sensing artifacts, and scene layouts in outdoor environments, which lead to unreliable spatial grounding and increased hallucination. (ii) current methods generally lack strong and transferable cross-modal alignment between 3D observations and language, resulting in limited OOD generalization across environments. Addressing these challenges requires not only improving in-domain spatial reasoning but also introducing principled mechanisms that enhance robustness under distribution shifts, particularly for indoor-to-outdoor transfer.

A promising direction is to leverage strong visual-linguistic priors from large-scale pre-training to improve semantic grounding and cross-modal alignment.
Foundation vision-language models such as CoCa~\citep{coca2022} demonstrate that contrastive pre-training on large image-text corpora yields representations with rich semantics and strong alignment between modalities.
Motivated by this, we develop 3D CoCa v2, a unified 3D captioning framework that combines contrastive vision-language learning with 3D caption generation in a shared feature space.
3D CoCa v2 builds on a frozen CLIP vision-language backbone for semantic priors, a spatially-aware 3D scene encoder for geometric context, and a multi-modal decoder that jointly optimizes contrastive and captioning objectives.
This unified design avoids reliance on external detectors or handcrafted proposals and establishes a strong captioner with improved semantic grounding.
However, even a strong unified captioner can still produce suboptimal outputs under domain shift, as standard decoding typically commits to a single caption without considering alternative hypotheses.
This motivates our key observation: \emph{test-time search over multiple candidates can improve robustness and faithfulness without updating model parameters}.

To this end, we introduce Test-Time Search for 3D CoCa v2.
As illustrated in the right panel of Fig.~\ref{fig:main}(a), TTS is an inference-only module built on top of the 3D CoCa \citep{huang20253dcoca} backbone; it generates multiple caption candidates and performs reward-guided selection conditioned on a compact scene summary.
By explicitly searching among plausible captions and selecting the one best supported by scene evidence, TTS serves as a simple plug-and-play mechanism that improves caption faithfulness under distribution shift, without additional training or parameter updates.
We evaluate 3D CoCa v2 in both in-domain and out-of-distribution settings.
On the indoor benchmarks ScanRefer~\citep{chen2020scanrefer} and Nr3D~\citep{achlioptas2020referit_3d}, 3D CoCa v2 consistently improves over 3D CoCa  by applying test-time search.
To assess cross-environment generalization, we further evaluate on the outdoor benchmark TOD$^3$Cap~\citep{todocap2025} in a zero-shot OOD setting.
As summarized in Fig.~\ref{fig:main}(b), 3D CoCa v2 achieves a further \textbf{+3.6} CIDEr@0.5 improvement over 3D CoCa, demonstrating stronger robustness under distribution shift.
In summary, the main contributions of this work include:
\begin{itemize}
\setlength\itemsep{0em}
    \item We present \textbf{3D CoCa v2}, a unified and end-to-end 3D captioning framework that combines contrastive vision-language learning with 3D caption generation in a shared feature space, avoiding external detectors or handcrafted proposals.
    \item We introduce Test-Time Search (TTS), a judge-guided reward-based inference strategy that generates diverse caption candidates and performs reward-guided selection using a compact scene summary, thereby improving robustness under domain shift without updating the captioner parameters.
    \item Extensive evaluations demonstrate that 3D CoCa v2 consistently improves over 3D CoCa \citep{huang20253dcoca} on the in-domain benchmarks ScanRefer and Nr3D. With the best-of-$N$ Test-Time Search ($N{=}8$ by default), 3D CoCa v2 achieves \textbf{+1.50} CIDEr@0.5 on ScanRefer (w/o additional 2D input) and \textbf{+1.61} CIDEr@0.5 on Nr3D. Moreover, under OOD evaluation on TOD$^3$Cap in a zero-shot setting, it yields a further \textbf{+3.6} CIDEr@0.5 gain over \citep{huang20253dcoca}, demonstrating improved robustness under distribution shift.
\end{itemize}

\begin{figure*}[t]
    \centering
    \includegraphics[width=0.9\linewidth]{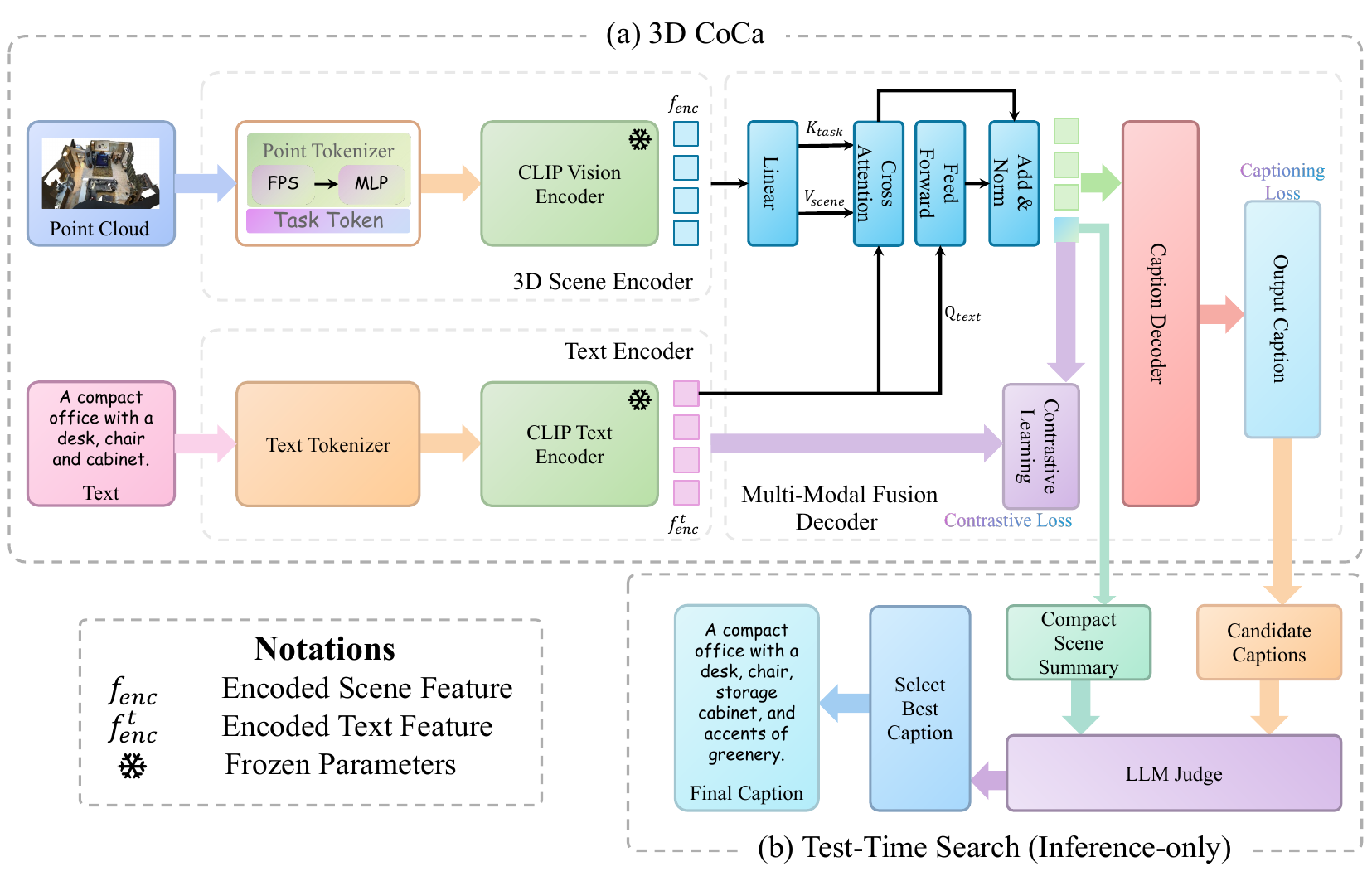}
    \caption{\textbf{Overview of 3D CoCa v2.}
    (a) \textbf{3D CoCa} learns aligned 3D–text representations by jointly optimizing contrastive alignment and caption generation: a point-cloud scene encoder and a text encoder produce fused features for a multi-modal decoder to generate a draft caption. (b) \textbf{Test-Time Search (inference-only)} improves robustness without any parameter updates by generating best-of-$N$ candidate captions from the backbone, conditioning an external LLM judge on a compact scene summary, and selecting the highest-scoring candidate as the final caption.}
    \label{fig:pipeline}
\end{figure*}

\section{Related Work}
\label{sec:related_work}
\noindent\textbf{3D captioning} localizes objects in 3D scenes and describes them in natural language.
As a high-level semantic interface, 3D captioning plays a crucial role in enabling spatial intelligence by requiring models to jointly perceive geometry, reason about spatial relationships, and communicate scene understanding through language.
Early works typically followed a two-stage ``detect-then-describe'' paradigm.
Scan2Cap~\citep{scan2cap_2021} pioneered this task by coupling 3D object localization with caption generation, and subsequent methods enhanced relational reasoning and contextual modeling, e.g., MORE~\citep{MORE_2022}.
Transformer-based architectures further accelerated progress.
SpaCap3D~\citep{spa2cap2022} employed an encoder-decoder design with spatially guided representations for geometry-aware captioning, while $\chi$-Trans2Cap~\citep{trans2cap2022} distilled semantic knowledge from 2D vision-language models into a 3D captioner.
Recent works also pursue unified multi-task formulations, such as 3DJCG~\citep{3djcg2022} and UniT3D~\citep{unit3d2023}, which jointly optimize captioning with related grounding or scene understanding objectives.
To mitigate error propagation from staged pipelines, end-to-end paradigms have been explored.
Vote2Cap-DETR~\citep{vote2cap2023} and Vote2Cap-DETR++~\citep{vote2cap++2024} reformulate dense captioning as a set-prediction problem and jointly localize and describe objects in a single forward pass.
TOD$^3$Cap~\citep{todocap2025} further targets outdoor environments and highlights the importance of robustness under domain shift, which remains challenging for indoor-trained 3D captioners.

\noindent\textbf{3D pre-training and vision-language models.}
Another line of research focuses on learning transferable 3D representations via pre-training.
Unsupervised 3D representation learning can be broadly grouped into global contrastive methods~\citep{Wang_Liu_Yue_Lasenby_Kusner_2021,Mei_Huang_Liu_Zhang_Wu_2022}, local contrastive objectives~\citep{PointContrast2020,wang2023tap}, and masked point modeling approaches~\citep{yu2021pointbert,pang2022masked}.
At a high level, contrastive learning \cite{cai2022dual,tang2020edge,zhuang2024mining} encourages transferable representations by pulling semantically similar samples together in the embedding space while pushing apart dissimilar ones, making it a natural choice for unsupervised and weakly supervised 3D pre-training.
These methods learn strong geometric features but do not explicitly ground 3D representations in natural language.
To bridge this gap, 3D vision-language pre-training aligns 3D regions or segments with text descriptions~\citep{huang2025dc}.
For example, 3D-VLP~\citep{3dvlp2024} aligns point cloud segments with language using contrastive learning, and UniT3D~\citep{unit3d2023} demonstrates that pre-training on large-scale point cloud–caption pairs benefits multiple 3D understanding tasks.
Beyond pre-training, recent 3D vision-language foundation models further improve 3D–language alignment and generalization through enhanced reasoning and instruction tuning.
For instance, 3D-R1~\citep{huang20253d} studies unified 3D reasoning across diverse scene understanding tasks, and its reasoning-oriented representations have also shown promise for downstream embodied settings that require grounded perception and decision making~\citep{huang2025mobilevla,liu2025evovla,ye2025vla,liu2025nav,song2025maniplvm,song2025hazards}.
Overall, these advances motivate unified contrastive–generative frameworks that strengthen cross-modal alignment and semantic grounding for 3D captioning.

\noindent\textbf{Test-time search and judging.}
Beyond training-time modeling, inference-time strategies have been studied to improve generation quality without updating model parameters.
A common approach is to generate multiple candidates and select the best output using an auxiliary scoring signal, including self-consistency style aggregation~\citep{wang2023selfconsistency} and related sampling-and-selection schemes~\citep{ichihara2025evaluation}.
Recent advances also explore large language models as judges to provide preference signals for open-ended evaluation and selection, such as MT-Bench and Chatbot Arena~\citep{zheng2023judging} and rubric-based frameworks like G-Eval~\citep{liu2023geval}.
In the agent setting, AgentRM~\citep{xia2025agentrm} investigates judge-guided search for improved generalization, and training-free pipelines employ LLM-driven components for open-world decision making.
These developments motivate using test-time search with external judging signals as a plug-and-play mechanism for robustness.
Different from prior text-only judging setups, 3D captioning requires bridging the modality gap between point clouds and language-based judges, which motivates compact scene summaries for reliable test-time selection.

\section{The Proposed Method}
\label{sec:method}

\subsection{Overview}
In this section, we present \textbf{3D CoCa v2}, a generalizable framework that bridges 3D point cloud representations and natural language for 3D captioning.
3D CoCa v2 follows a unified contrastive-generative design, drawing inspiration from CLIP-style vision-language pre-training~\citep{clip2021} and the Contrastive Captioner (CoCa) paradigm~\citep{coca2022}.
As illustrated in Fig.~\ref{fig:pipeline}(a), the backbone consists of four components: a \textit{3D Scene Encoder}, a \textit{Text Encoder}, a \textit{Contrastive Learning} module, and a \textit{Multi-Modal Fusion Decoder}.

Compared with 3D CoCa, the key extension of 3D CoCa v2 is an inference-time \emph{Test-Time Search} procedure, as illustrated in Fig.~\ref{fig:main}(a) and Fig.~\ref{fig:pipeline}(b), which improves robustness under distribution shifts \emph{without updating the captioner parameters}.
Instead of producing a single caption by standard decoding, TTS generates a set of diverse candidates and performs reward-guided selection using a compact scene summary.
In our setting, the reward is provided by an external large language model acting as a judge.
The backbone is trained end-to-end with contrastive and captioning objectives, while TTS is applied only at inference time and requires no additional training.

\subsection{3D Scene Encoder}
\label{subsec:sceneencoder}
The 3D scene encoder transforms an unstructured point cloud into a set of latent tokens that capture geometric and semantic content.
It integrates point-based 3D processing with a frozen 2D CLIP visual backbone to capture both geometry and semantics.
It comprises three components: (i) a point cloud tokenizer that partitions raw point clouds into patch tokens, (ii) learnable task tokens that inject 3D captioning context, and (iii) a frozen CLIP Vision Transformer that encodes the concatenated token sequence.
As shown in Fig.~\ref{fig:pipeline} (top-left), the encoder converts the raw 3D input into a structured representation suitable for multimodal reasoning.

\noindent\textbf{Point cloud tokenizer.}
Given an input point cloud $P \in \mathbb{R}^{N \times (3+F)}$, each point is described by 3D coordinates $(x,y,z)$ and $F$ additional features (e.g., color or normals).
We convert the point cloud into a discrete token sequence for transformer processing.
We use farthest point sampling (FPS) to select $M$ representative points as patch centers.
For each center, we gather its $K$ nearest neighbors to form a local patch, producing $M$ patches $\{P_1,\dots,P_M\}$, each containing $K$ points.
Each patch is encoded by a lightweight point-wise encoder implemented as multi-layer perceptrons (MLPs), yielding $M$ point tokens of dimension $D_p$:
\begin{equation}
E_p(P) = [\mathbf{e}_{p_1}, \mathbf{e}_{p_2}, \dots, \mathbf{e}_{p_M}] \in \mathbb{R}^{M \times D_p},
\end{equation}
where $\mathbf{e}_{p_i}$ denotes the embedding of the $i$-th patch.

\noindent\textbf{Task token.}
Point tokens capture local geometry and appearance but lack explicit task awareness.
To guide the model toward captioning, we introduce $m_t$ learnable task tokens that are prepended to the point token sequence.
Inspired by prompt tuning~\citep{ptuning2022}, task tokens act as high-level prompts that aggregate global semantic cues (e.g., layout and salient objects) via self-attention and condition the encoder for language-relevant feature extraction.

\noindent\textbf{Frozen CLIP vision encoder.}
We concatenate the point tokens and task tokens into a unified sequence:
\begin{equation}
[\mathbf{e}_{p_1}, \dots, \mathbf{e}_{p_M}; \mathbf{t}_1, \dots, \mathbf{t}_{m_t}],
\end{equation}
where $\mathbf{t}_j$ denotes the $j$-th task token.
This sequence is fed into a frozen CLIP Vision Transformer~\citep{clip2021}.
All CLIP weights are kept frozen to preserve pre-trained visual representations and stabilize optimization.
The CLIP encoder outputs latent embeddings that jointly capture 3D geometry and task context.
We extract a global scene representation $f_{\text{enc}} \in \mathbb{R}^{D}$ as the scene embedding used for contrastive alignment and captioning.

\subsection{Text Encoder}
The text encoder maps natural language descriptions into a semantically aligned embedding space.
We adopt the Transformer-based CLIP text encoder~\citep{clip2021} and keep its weights frozen to retain the rich linguistic knowledge acquired during large-scale pretraining.

\noindent\textbf{Text tokenizer.}
Given an input caption $T$, we tokenize it into $L$ subword tokens and map them to embeddings:
\begin{equation}
E_t(T) = [\mathbf{e}_{t_1}, \mathbf{e}_{t_2}, \dots, \mathbf{e}_{t_L}] \in \mathbb{R}^{L \times D_t},
\end{equation}
where $\mathbf{e}_{t_i}$ is the embedding of the $i$-th token.
We add positional encodings and prepend a special token used as a sentence-level aggregator.

\noindent\textbf{Frozen CLIP text encoder.}
The token embeddings are processed by the CLIP text Transformer, consisting of $N_{\text{te}}$ layers:
\begin{equation}
\resizebox{0.9\hsize}{!}{$
H^l = \mathrm{TransformerBlock}^l(H^{l-1}), \qquad l \in [1,\dots,N_{\text{te}}],
$}
\end{equation}
with $H^0 = E_t(T)$.
All weights are frozen.
We take the hidden state of the special token from the final layer as the global text representation
$f_{\text{enc}}^t \in \mathbb{R}^{D_t}$,
which serves as the language-side embedding in contrastive learning.

\subsection{Contrastive Learning}
To align 3D scenes and text, we employ a contrastive objective that maps the scene feature $f_{\text{enc}}$ and the text feature $f_{\text{enc}}^t$ into a shared embedding space.
Matched 3D-text pairs are brought together, while mismatched pairs are pushed apart, following the CLIP paradigm.

\noindent\textbf{Feature alignment.}
We project both features into a shared space using learnable projection heads:
\begin{equation}
\tilde{f}_{\text{enc}}=\mathrm{MLP}_v(f_{\text{enc}}),\qquad
\tilde{f}_{\text{enc}}^t=\mathrm{MLP}_t(f_{\text{enc}}^t),
\end{equation}
and apply L2 normalization:
\begin{equation}
\hat{f}_{\text{enc}}=\frac{\tilde{f}_{\text{enc}}}{\|\tilde{f}_{\text{enc}}\|_2},\qquad
\hat{f}_{\text{enc}}^t=\frac{\tilde{f}_{\text{enc}}^t}{\|\tilde{f}_{\text{enc}}^t\|_2}.
\end{equation}

\noindent\textbf{Contrastive loss.}
For a batch of $N$ paired samples, the cosine similarity between scene $i$ and text $j$ is
\begin{equation}
    \mathrm{sim}\left ( \hat{f}_{\text{enc},i};\hat{f}_{\text{enc},j}^t   \right )  =\frac{\hat{f}_{\text{enc},i}\cdot \hat{f}_{\text{enc},j}^t  }{\left \| \hat{f}_{\text{enc},i}  \right \| \left \| \hat{f}_{\text{enc},j}^t  \right \|}, 
\end{equation}
We use an InfoNCE loss:
\begin{equation}
\label{eq:losscon}
\resizebox{0.9\hsize}{!}{$
\mathcal{L}_{\mathrm{Con}} =
-\frac{1}{N}\sum_{i=1}^{N}
\log
\frac{\exp\bigl(\mathrm{sim}(\hat{f}_{\text{enc},i};\hat{f}_{\text{enc},i}^t)/\tau\bigr)}
{\sum_{j=1}^{N}\exp\bigl(\mathrm{sim}(\hat{f}_{\text{enc},i};\hat{f}_{\text{enc},j}^t)/\tau\bigr)},
$}
\end{equation}
where $\tau$ is a learnable temperature.

\subsection{Multi-Modal Fusion Decoder}
The multi-modal fusion decoder generates captions conditioned on the input 3D scene.
It is implemented as an autoregressive Transformer decoder with cross-attention, generating tokens one by one.
At time step $t$, the decoder predicts $y_t$ conditioned on the previously generated tokens $y_{<t}$ via causal self-attention and on the encoded scene representation via cross-attention:
$P(y_t \mid y_{<t}, f_{\text{enc}})$.

\begin{algorithm}[t]\small
\caption{Training of 3D CoCa v2}
\label{alg:train}
\begin{algorithmic}[1]
\REQUIRE Point cloud $P$, paired caption $T$
\STATE $\mathbf{E}_p \leftarrow \mathrm{Tokenizer}_{3D}(P)$
\STATE $\mathbf{E}_t \leftarrow \mathrm{Tokenizer}_{text}(T)$
\STATE $\mathbf{f}_{enc} \leftarrow \mathrm{CLIP}_{vis}(\mathbf{E}_p)$ \COMMENT{frozen}
\STATE $\mathbf{f}_{enc}^{t} \leftarrow \mathrm{CLIP}_{txt}(\mathbf{E}_t)$ \COMMENT{frozen}
\STATE $\hat{\mathbf{f}}_{enc},\hat{\mathbf{f}}_{enc}^{t} \leftarrow \mathrm{ProjNorm}(\mathbf{f}_{enc},\mathbf{f}_{enc}^{t})$
\STATE $\mathcal{L}_{Con} \leftarrow \mathrm{InfoNCE}(\hat{\mathbf{f}}_{enc},\hat{\mathbf{f}}_{enc}^{t})$
\STATE $\hat{C} \leftarrow \mathrm{Decoder}(\mathbf{f}_{enc})$
\STATE $\mathcal{L}_{Cap} \leftarrow \mathrm{CE}(\hat{C}, T)$
\STATE $\mathcal{L}_{Total} \leftarrow \mathcal{L}_{Con} + \lambda \mathcal{L}_{Cap}$
\STATE Update trainable parameters with $\nabla \mathcal{L}_{Total}$
\end{algorithmic}
\end{algorithm}

\noindent\textbf{Cross-attention mechanism.}
Let $Q_{\text{text}}$ be the query matrix from the decoder hidden states, and let $K_{\text{scene}}, V_{\text{scene}}$ be the key and value matrices derived from scene tokens.
Cross-attention is computed as
\begin{equation}
\label{eq:crossattn}
\resizebox{0.9\hsize}{!}{$
\mathrm{Attention}(Q_{\text{text}},K_{\text{scene}},V_{\text{scene}})
=
\mathrm{softmax}\Bigl(\frac{Q_{\text{text}}K_{\text{scene}}^{\top}}{\sqrt{d_k}}\Bigr)V_{\text{scene}},
$}
\end{equation}
where $d_k$ is the key dimensionality.

\subsection{Training Objectives and Joint Optimization}
\label{sec:train}
We train the backbone with a combination of contrastive loss and captioning loss.
The overall backbone training procedure is summarized in Alg.~\ref{alg:train}.
The contrastive loss $\mathcal{L}_{\mathrm{Con}}$ in Eq.~\eqref{eq:losscon} aligns scene and text features in a shared space.
The decoder is supervised with a standard cross-entropy captioning loss.
Given a predicted caption $\hat{Y} = (\hat{y}_1, \dots, \hat{y}_L)$ and the corresponding ground-truth sequence $Y = (y_1, \dots, y_L)$, the captioning loss is defined as:
\begin{equation}
    \mathcal{L}_{\mathrm{Cap}}=-\sum_{t=1}^{L} \log P\left ( \hat{y}_t=y_t\mid \hat{y}_{< t} ,f_{\text{enc}}  \right ) ,
\end{equation}
where $f_{\text{enc}}$ is the global 3D scene embedding used to condition the decoder via cross-attention.

The overall objective is
\begin{equation}
\mathcal{L}_{\mathrm{Total}} = \mathcal{L}_{\mathrm{Con}} + \lambda \cdot \mathcal{L}_{\mathrm{Cap}},
\end{equation}
where $\lambda$ balances alignment and generation.
Notably, the proposed Test-Time Search is applied only during inference and does not introduce additional trainable parameters or losses in backbone optimization.

\subsection{Test-Time Search}
\label{subsec:tts}
Standard decoding commits to a single caption and may be brittle under distribution shifts.
To improve faithfulness without updating model parameters, 3D CoCa v2 introduces \emph{Test-Time Search}, which performs reward-guided selection over multiple candidate captions.
The full inference-time procedure is summarized in Alg.~\ref{alg:tts}.

\noindent\textbf{Compact scene summary.}
A language-only judge cannot directly interpret raw point clouds.
TTS conditions the judge on a compact scene summary $S(P)$ derived from the scene embedding.
We construct $S(P)$ via retrieval in the shared contrastive space.
Let $\mathcal{B}=\{b_k\}$ be a bank of short textual descriptors encoded by the frozen CLIP text encoder, and let $\phi(\cdot)$ denote their embeddings.
Given $\hat{f}_{\text{enc}}$, we retrieve the top-$K_s$ descriptors and concatenate them into a compact summary:
\begin{equation}
S(P) = \mathrm{Concat}\Bigl(\mathrm{TopK}(\mathrm{sim}(\hat{f}_{\text{enc}}, \phi(b_k)))\Bigr).
\end{equation}
This summary provides high-level evidence, such as salient objects and scene types.
Alternatively, $S(P)$ can be produced by structured decoding; we adopt retrieval for simplicity and reproducibility.

\begin{algorithm}[t]\small
\caption{TTS for 3D CoCa v2}
\label{alg:tts}
\begin{algorithmic}[1]
\REQUIRE Point cloud $P$, candidate size $N$, summary size $K_s$
\ENSURE Final caption $C^{*}$
\STATE $\mathbf{f}_{enc} \leftarrow \mathrm{CLIP}_{vis}(\mathrm{Tokenizer}_{3D}(P))$ \COMMENT{frozen backbone}
\STATE $\hat{\mathbf{f}}_{enc} \leftarrow \mathrm{ProjNorm}(\mathbf{f}_{enc})$
\STATE $S(P) \leftarrow \mathrm{RetrieveSummary}(\hat{\mathbf{f}}_{enc}, K_s)$
\STATE Generate $\{C_i\}_{i=1}^{N}$ by stochastic decoding from $\mathrm{Decoder}(\mathbf{f}_{enc})$
\FOR{$i=1$ to $N$}
    \STATE $r_i \leftarrow J(S(P), C_i)$ \COMMENT{external judge}
\ENDFOR
\STATE $C^{*} \leftarrow \arg\max_{C_i} r_i$
\end{algorithmic}
\end{algorithm}

\noindent\textbf{Candidate generation and selection.}
Given $P$, we generate a set of $N$ diverse candidates $\{C_i\}_{i=1}^{N}$ using stochastic decoding.
An external large language model acts as a judge and outputs a scalar reward for each candidate conditioned on $S(P)$:
\begin{equation}
r_i = J(S(P), C_i).
\end{equation}
The final caption is selected by
\begin{equation}
C^{*} = \arg\max_{C_i} r_i,
\end{equation}
and top-$k$ voting can be applied when multiple high-scoring candidates are retained.
All prompts and scoring rubrics used for the judges are fixed across datasets and are reported for reproducibility.

\begin{table*}
\begin{center}
    \caption{
    \textbf{Comparison on ScanRefer~\cite{chen2020scanrefer}.} We evaluate the performance of each method, with and without additional 2D input, at IoU thresholds of 0.25 and 0.5. Metrics include CIDEr (C)~\citep{cider2015}, BLEU-4 (B-4)~\citep{bleu2002}, METEOR (M)~\citep{meteor2005}, and ROUGE-L (R)~\citep{rouge2004}.
    }
    \resizebox{\linewidth}{!}{
    \begin{tabular}{lccccccccccccccccccc}
    \toprule
    \multirow{3}{*}{Method} & \multicolumn{9}{c}{w/o additional 2D input} &  & \multicolumn{9}{c}{w/ additional 2D input} \\
    \cline{2-10}\cline{12-20}
    & \multicolumn{4}{c}{IoU = 0.25} &  & \multicolumn{4}{c}{IoU = 0.50} &  & \multicolumn{4}{c}{IoU = 0.25} &  & \multicolumn{4}{c}{IoU = 0.50} \\
    \cline{2-5} \cline{7-10} \cline{12-15} \cline{17-20} 
    & C$\uparrow$ & B-4$\uparrow$ & M$\uparrow$ & R$\uparrow$ &  & C$\uparrow$ & B-4$\uparrow$ & M$\uparrow$ & R$\uparrow$ &  & C$\uparrow$ & B-4$\uparrow$ & M$\uparrow$ & R$\uparrow$ &  & C$\uparrow$ & B-4$\uparrow$ & M$\uparrow$ & R$\uparrow$ \\ \hline
    Scan2Cap~\citep{scan2cap_2021}& 53.73 & 34.25 & 26.14 & 54.95 & & 35.20 & 22.36 & 21.44 & 43.57 & & 56.82 & 34.18 & 26.29 & 55.27 & & 39.08 & 23.32 & 21.97& 44.78 \\
    MORE~\citep{MORE_2022}        & 58.89 & 35.41 & 26.36 & 55.41 & & 38.98 & 23.01 & 21.65 & 44.33 & & 62.91 & 36.25 & 26.75 & 56.33 & & 40.94 & 22.93 & 21.66& 44.42 \\
    SpaCap3d~\citep{spa2cap2022}  & 58.06 & 35.30 & 26.16 & 55.03 & & 42.76 & 25.38 & 22.84 & 45.66 & & 63.30 & 36.46 & 26.71 & 55.71 & & 44.02 & 25.26 & 22.33& 45.36 \\
    3DJCG~\citep{3djcg2022}       & 60.86 & 39.67 & 27.45 & 59.02 & & 47.68 & 31.53 & 24.28 & 51.80 & & 64.70 & 40.17 & 27.66 & 59.23 & & 49.48 & 31.03 & 24.22& 50.80 \\
    D3Net~\citep{chen2021d3net}   & -     & -     & -     & -     & & -     & -     & -     & -     & & -     & -     & -     & -     & & 46.07 & 30.29 & 24.35& 51.67 \\
    3D-VLP~\citep{3dvlp2024}      & 64.09 & 39.84 & 27.65 & 58.78 & & 50.02 & 31.87 & 24.53 & 51.17 & & 70.73 & 41.03 & 28.14 & 59.72 & & 54.94 & 32.31 & 24.83& 51.51 \\
    Vote2Cap-DETR~\citep{vote2cap2023}& 71.45 & 39.34 & 28.25 & 59.33 & & 61.81 & 34.46 & 26.22 & 54.40 & & 72.79 & 39.17 & 28.06 & 59.23 & & 59.32 & 32.42& 25.28 & 52.53 \\
    Unit3D~\citep{unit3d2023}     & -     & -     & -     & -     & & -  & -  & - & - &  & - & - & - & - & & 46.69 & 27.22 & 21.91 & 45.98 \\
    Vote2Cap-DETR++~\citep{vote2cap++2024}& 76.36 & 41.37 & 28.70 & 60.00 & & 67.58 & 37.05 & 26.89 & 55.64 & & 77.03 & 40.99 & 28.53 & 59.59 & & 64.32 & 34.73& 26.04 & 53.67 \\
    BiCA~\citep{bica2025}         & 78.42 & 41.46 & 28.82 & 60.02 & &68.46 & 38.23 & 27.56 & 58.56 & & 78.35 & 41.20 & 28.82 & 59.80 & & 66.47 & 36.13 & 26.71 & 54.54 \\
    See-It-All~\citep{seeitall2024}& 78.68 & 43.25 & 29.21 & 63.06 & & 73.22 & 40.91 & 28.19 & 60.46 & & 78.05 & 42.16 & 28.74 & 61.70 & & 69.86 & 37.89 & 27.04 & \textbf{57.33} \\
    \midrule
    3D CoCa~\citep{huang20253dcoca} & 85.42 & 45.56 & 30.95 & 61.98 & & 77.13 & 41.23 & 28.52& 57.40& & 86.12 & 44.79 & 30.75& 61.45  & & 74.52 & 38.42 & 28.03 & 55.23 \\
    \textbf{3D CoCa v2 (Ours)} & \textbf{86.95} & \textbf{45.99} & \textbf{31.55} & \textbf{63.98} & & \textbf{78.63} & \textbf{41.55} & \textbf{28.95}& \textbf{61.95} & & \textbf{87.05} & \textbf{45.10} & \textbf{31.78}& \textbf{62.45}  & & \textbf{75.60} & \textbf{38.70} & \textbf{28.25} & 55.63       \\
     & \textbf{\textcolor{green!60!black}{+1.53}} & \textbf{\textcolor{green!60!black}{+0.43}} & \textbf{\textcolor{green!60!black}{+0.60}} & \textbf{\textcolor{green!60!black}{+0.92}} & & \textbf{\textcolor{green!60!black}{+1.50}} & \textbf{\textcolor{green!60!black}{+0.32}} & \textbf{\textcolor{green!60!black}{+0.43}} & \textbf{\textcolor{green!60!black}{+1.49}} & & \textbf{\textcolor{green!60!black}{+0.93}} & \textbf{\textcolor{green!60!black}{+0.31}} & \textbf{\textcolor{green!60!black}{+1.03}} & \textbf{\textcolor{green!60!black}{+0.75}}& &\textbf{\textcolor{green!60!black}{+1.08}} & \textbf{\textcolor{green!60!black}{+0.38}}&\textbf{\textcolor{green!60!black}{+0.22}}& \textbf{\textcolor{green!60!black}{+0.4}} \\
    \bottomrule
    \end{tabular}
    }
    \label{exp:comparison_on_scanrefer}
    \vspace{-0.4cm}
\end{center}
\end{table*}

\begin{table}[t]
    \centering
    \caption{\textbf{Results on Nr3D~\cite{achlioptas2020referit_3d} at IoU=0.5.} We report CIDEr (C)~\cite{cider2015}, BLEU-4 (B-4)~\cite{bleu2002}, METEOR (M)~\cite{meteor2005}, and ROUGE-L (R)~\cite{rouge2004}.}
    \resizebox{\linewidth}{!}{
    \begin{tabular}{lcccc}
    \toprule
    Method & C@0.5$\uparrow$ & B-4@0.5$\uparrow$ & M@0.5$\uparrow$ & R@0.5$\uparrow$ \\ \hline
    Scan2Cap~\citep{scan2cap_2021} & 27.47 & 17.24 & 21.80 & 49.06 \\
    SpaCap3d~\citep{spa2cap2022}   & 33.71 & 19.92 & 22.61 & 50.50 \\
    D3Net~\citep{chen2021d3net}    & 33.85 & 20.70 & 23.13 & 53.38 \\
    3DJCG~\citep{3djcg2022}        & 38.06 & 22.82 & 23.77 & 52.99 \\
    Vote2Cap-DETR~\citep{vote2cap2023} & 43.84 & 26.68 & 25.41 & 54.43 \\
    Vote2Cap-DETR++~\citep{vote2cap++2024} & 47.08 & 27.70 & 25.44 & 55.22 \\
    BiCA~\citep{bica2025}          & 48.77 & 28.35 & 25.60 & 55.81 \\
    \midrule
    3D CoCa~\citep{huang20253dcoca} & 52.84  & 29.29 & 25.55 & 56.43  \\
    \textbf{3D CoCa v2 (Ours)}             & \textbf{54.45}  & \textbf{29.85} & \textbf{25.95} & \textbf{57.12}  \\
               & \textbf{\textcolor{green!60!black}{+1.61}} & \textbf{\textcolor{green!60!black}{+0.56}} & \textbf{\textcolor{green!60!black}{+0.35}} & \textbf{\textcolor{green!60!black}{+0.69}} \\
    \bottomrule
    \end{tabular}
    }
    \label{exp:comparison_on_nr3d}
    \vspace{-0.4cm}
\end{table}

\section{Experiments}
\subsection{Datasets and Evaluation Metrics}
\noindent\textbf{Datasets.}
We evaluate \emph{in-domain} 3D captioning on ScanRefer~\citep{chen2020scanrefer} and Nr3D~\citep{achlioptas2020referit_3d}, which provide human-annotated descriptions for objects in indoor 3D scenes.
ScanRefer contains 36,665 descriptions for 7,875 objects across 562 scenes, while Nr3D includes 32,919 descriptions for 4,664 objects in 511 scenes.
Both benchmarks are derived from ScanNet~\citep{dai2017scannet}, which comprises 1,201 reconstructed indoor scenes.
We follow the standard validation splits used in prior work: ScanRefer includes 9,508 descriptions for 2,068 objects in 141 scenes, and Nr3D includes 8,584 descriptions for 1,214 objects in 130 scenes.
All evaluation scenes are drawn from the ScanNet validation set, ensuring a consistent protocol across benchmarks.

To assess OOD generalization across environments, we further evaluate on the outdoor 3D dense captioning benchmark TOD$^3$Cap~\citep{todocap2025}.
Since our focus is on caption generalization rather than detection, we adopt an \emph{oracle-box} setting on TOD$^3$Cap, where ground-truth 3D boxes are provided at test time to isolate the captioning component under distribution shift.

\noindent\textbf{Evaluation metrics.}
We report standard captioning metrics, including CIDEr~\citep{cider2015}, BLEU-4~\citep{bleu2002}, METEOR~\citep{meteor2005}, and ROUGE-L~\citep{rouge2004}, denoted as C, B-4, M, and R, respectively.
Following common practice in 3D dense captioning~\citep{scan2cap_2021,MORE_2022,spa2cap2022,vote2cap2023,3djcg2022}, we evaluate caption quality under the localization-aware $m@k$IoU protocol~\citep{scan2cap_2021}.
Given $N$ annotated objects, the metric is defined as:
\begin{equation}
\label{eq:mk_iou}
\resizebox{0.85\hsize}{!}{$
m@k\text{IoU}=\frac{1}{N}\sum_{i=1}^{N} m\!\left(\hat{c}_i,c_i\right)\cdot
\mathbb{I}\!\left\{\text{IoU}\!\left(\hat{b}_i,b_i\right)\ge k\right\},
$}
\end{equation}
where $\hat{c}_i$ and $c_i$ denote the predicted and ground-truth captions, $\hat{b}_i$ and $b_i$ are the predicted and ground-truth 3D bounding boxes, and $m(\cdot)$ is a captioning metric (e.g., CIDEr, METEOR, BLEU-4, ROUGE-L).
For ScanRefer and Nr3D, we report results at the standard IoU thresholds (0.25 and 0.5 for ScanRefer and 0.5 for Nr3D).

We evaluate on TOD$^3$Cap following the benchmark's standard captioning metrics and IoU thresholds and report CIDEr at IoU $\in\{0.25,0.5\}$ for consistency with prior work.
Since our goal is to assess caption \emph{generalization} rather than detection, we adopt an \emph{oracle-box} setting on TOD$^3$Cap, where ground-truth 3D boxes are provided at test time (i.e., $\hat{b}_i=b_i$).
Therefore, the IoU-gating indicator in Eq.~\eqref{eq:mk_iou} is always satisfied, and the reported scores reflect caption quality conditioned on correct localization.

To quantify faithfulness under test-time search, we additionally report a hallucination rate measuring the fraction of generated captions that mention objects or attributes not supported by scene evidence.
We compute this metric using the same verifier for all methods

\begin{table*}[!ht]
\caption{
\textbf{Results on TOD$^3$Cap~\citep{todocap2025} under two protocols.}
\textbf{Top:} in-domain training on TOD$^3$Cap. \textbf{Bottom:} zero-shot OOD evaluation on TOD$^3$Cap (trained on indoor data only, no TOD$^3$Cap fine-tuning). “$^*$” indicates replacing the scene encoder with a BEV encoder for adaptation, following~\citep{todocap2025}.
}
\label{tab:tod3cap_two_protocols}
\centering
\resizebox{\linewidth}{!}{
\begin{tabular}{l|c|cccc|cccc}
\toprule
Method & Venue & C@0.25$\uparrow$ & B-4@0.25$\uparrow$ & M@0.25$\uparrow$ & R@0.25$\uparrow$ & C@0.5$\uparrow$ & B-4@0.5$\uparrow$ & M@0.5$\uparrow$ & R@0.5$\uparrow$ \\
\midrule

\multicolumn{10}{c}{\textbf{In-domain (train on TOD$^3$Cap)}} \\
\midrule
Scan2Cap$^*$~\citep{scan2cap_2021}      & CVPR 2021 & 50.6 & 34.3 & 25.2 & 57.9 & 43.3 & 31.3 & 22.8 & 50.8 \\
Vote2Cap-DETR$^*$~\citep{vote2cap2023}  & CVPR 2023 & 72.8 & 41.6 & 29.5 & 60.6 & 62.6 & 35.9 & 27.4 & 55.8 \\
Vote2Cap-DETR++$^*$~\citep{vote2cap++2024}& T-PAMI 2024 & 78.8 & 42.6 & 29.9 & 60.8 & 66.5 & 36.9 & 27.4 & 56.2 \\
$\text{TOD}^3$Cap~\citep{todocap2025}   & ECCV 2024 & 85.3 & 43.0 & 29.9 & 60.5 & 74.4 & 39.4 & 27.2 & 55.4 \\

\midrule
\multicolumn{10}{c}{\textbf{OOD generalization (zero-shot on TOD$^3$Cap)}} \\
\midrule
Scan2Cap~\citep{scan2cap_2021}        & CVPR 2021 & 42.6 & 23.1 & 20.8 & 48.3 & 36.4 & 20.4 & 19.4 & 45.1 \\
Vote2Cap-DERT~\citep{vote2cap2023}    & CVPR 2023 & 49.8 & 27.0 & 22.6 & 51.2 & 43.2 & 24.2 & 21.2 & 48.7 \\
Vote2Cap-DETR++~\citep{vote2cap++2024}& T-PAMI 2024 & 52.1 & 28.1 & 23.1 & 52.0 & 45.4 & 25.0 & 21.5 & 49.4 \\
3D-VLP~\citep{3dvlp2024}              & AAAI 2024 & 52.9 & 28.7 & 23.4 & 52.6 & 45.9 & 25.4 & 21.7 & 49.8 \\ \midrule
3D CoCa~\citep{huang20253dcoca} & 3DV 2026 & 55.8 & 30.5 & 24.0 & 54.2 & 47.2 & 26.8 & 22.1 & 50.1 \\
\textbf{3D CoCa v2 (Ours)} & - & \textbf{59.6} & \textbf{31.6} & \textbf{24.6} & \textbf{55.7} &
\textbf{50.8} & \textbf{27.6} & \textbf{22.6} & \textbf{51.3} \\
 & - & \textbf{\textcolor{green!60!black}{+3.8}} & \textbf{\textcolor{green!60!black}{+1.1}} & \textbf{\textcolor{green!60!black}{+0.6}} & \textbf{\textcolor{green!60!black}{+1.5}} & \textbf{\textcolor{green!60!black}{+3.6}} & \textbf{\textcolor{green!60!black}{+0.8}} & \textbf{\textcolor{green!60!black}{+0.5}} & \textbf{\textcolor{green!60!black}{+1.2}} \\
\bottomrule
\end{tabular}
}
\vspace{-0.4cm}
\end{table*}

\subsection{Implementation Details}

\noindent\textbf{Input representation.}
The input point cloud $P \in \mathbb{R}^{N \times (3+F)}$ contains $N={40{,}000}$ points.
Each point includes its 3D coordinates and additional features.
In the \emph{w/o additional 2D} setting, we use color, normal, and height as per-point features.
In the \emph{w/ additional 2D} setting, we replace the raw color with 2D multi-view features extracted using ENet~\citep{enet2020}, following the established protocol in~\citep{scan2cap_2021}.

\noindent\textbf{Backbone training.}
We train the 3D CoCa v2 backbone with the joint contrastive and captioning objective described in sec.~\ref{sec:train} (Alg.~\ref{alg:train}).
Optimization uses AdamW~\citep{loshchilov2018decoupled} with learning rate ${\eta}={0.1}$, batch size ${B}={4}$, and a cosine annealing schedule.
Models are trained for ${E}={1080}$ epochs on ScanRefer and Nr3D.
All experiments are conducted on 2 $\times$ NVIDIA RTX 4090 GPUs.

\noindent\textbf{Test-Time Search (TTS).}
TTS is applied only at inference time and does not update backbone parameters (Alg.~\ref{alg:tts}).
For each scene, we sample ${N}={\mathbf{8}}$ candidate captions using stochastic decoding.
We construct a compact scene summary $S(P)$ via retrieval in the frozen contrastive space: we pre-encode a bank of short textual descriptors with the frozen CLIP text encoder and retrieve the top-${K_s}$ descriptors by cosine similarity to the scene embedding.
An external large language model acts as a judge and assigns a scalar reward to each candidate given $S(P)$ and the candidate caption.
The final caption is selected by maximum reward, with optional top-$k$ voting when multiple high-scoring candidates are retained.

\subsection{In-Domain Evaluation}
We compare 3D CoCa v2 against representative 3D dense captioning methods on ScanRefer and Nr3D using C, M, B-4, and R.
We report results under IoU thresholds of 0.25 and 0.5 on ScanRefer, and 0.5 on Nr3D, following prior work.
To highlight the contribution of inference-time optimization, we report two decoding settings for our method:
(i) \emph{standard decoding} using greedy or beam search, and
(ii) \emph{Test-Time Search (TTS)} with best-of-$N$ candidate selection.
Unless otherwise specified, we use best-of-$N$ TTS with $N=K$ candidates and set $K=8$ as a default trade-off between quality and inference cost.
TTS is an inference-only add-on. When TTS is disabled, the model reduces to the same contrastive-generative backbone as our 3D CoCa baseline, evaluated with standard decoding.

\begin{figure*}[htbp]
    \centering
    \includegraphics[width=\linewidth]{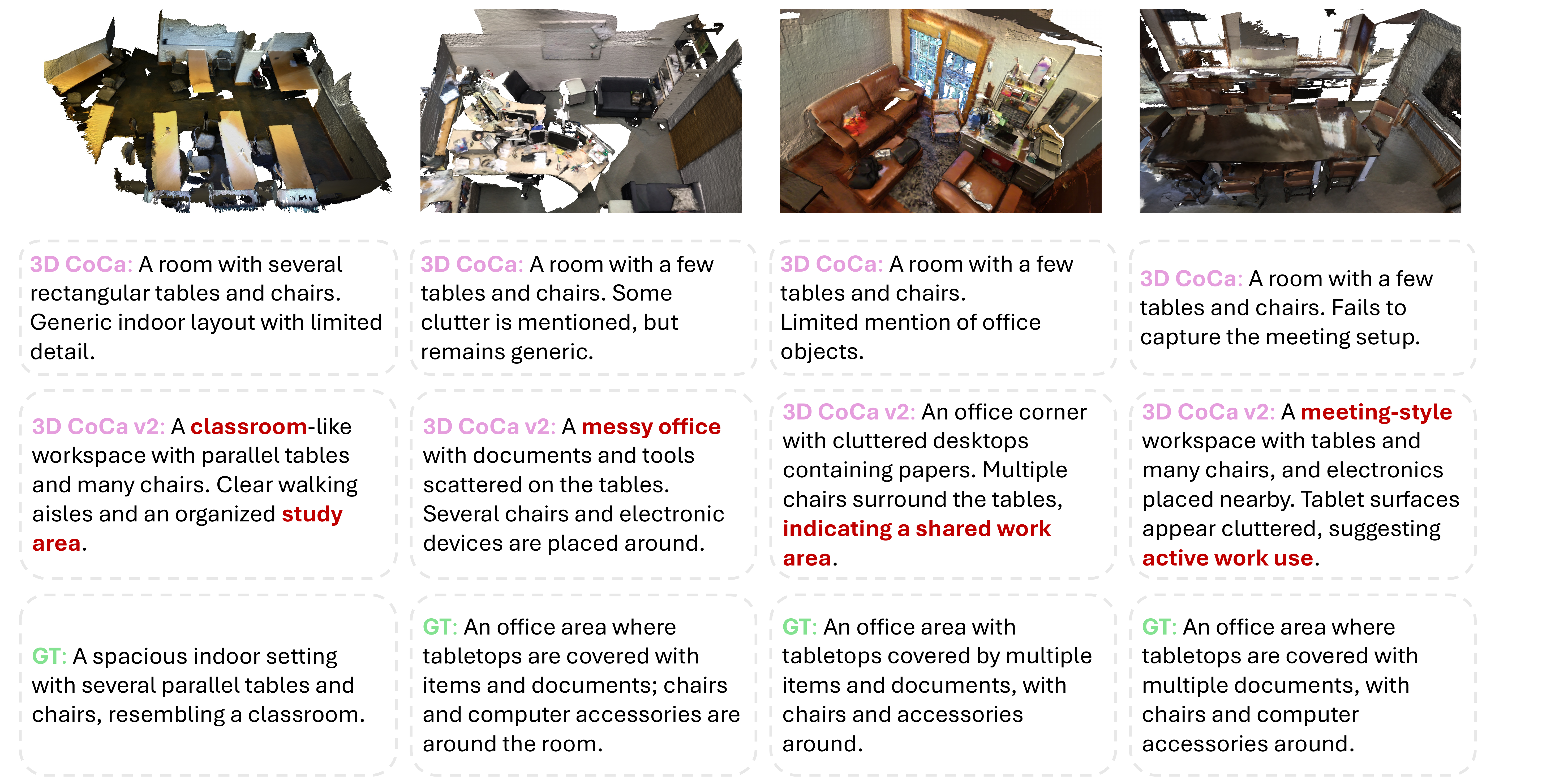}
    \caption{\textbf{Qualitative comparisons on ScanRefer~\citep{chen2020scanrefer}.}
    We visualize four representative scenes and the captions generated by 3D CoCa, 3D CoCa v2, and the ground truth (GT).
    Compared to the baseline, 3D CoCa v2 produces more detailed and better-grounded descriptions, capturing richer scene semantics and functional cues.
    Red-highlighted phrases mark the additional informative content provided by our method beyond the baseline.
    }
    \label{fig:visualization}
    \vspace{-0.4cm}
\end{figure*}

\noindent\textbf{ScanRefer.}
\tableautorefname~\ref{exp:comparison_on_scanrefer} shows that 3D CoCa v2 improves over 3D CoCa across both IoU thresholds and input settings.
Without additional 2D input, 3D CoCa v2 raises CIDEr from 85.42 to 86.95 at IoU=0.25 and from 77.13 to 78.63 at IoU=0.50, with consistent gains on BLEU-4, METEOR and ROUGE-L.
Overall, these results indicate that 3D CoCa v2 produces more informative captions while maintaining strong localization-aware caption quality under the standard ScanRefer protocol.

\noindent\textbf{Nr3D.}
On Nr3D (\tableautorefname~\ref{exp:comparison_on_nr3d}), 3D CoCa v2 achieves consistent improvements over prior methods at IoU$=0.5$.
These gains indicate stronger semantic grounding for the free-form referring expressions in Nr3D, where captions must capture fine-grained attributes and contextual cues beyond object categories.

\subsection{Out-of-Domain Evaluation}

\noindent\textbf{TOD$^3$Cap.}
We evaluate cross-environment generalization on TOD$^3$Cap using a zero-shot OOD protocol: all models are trained on indoor data only and evaluated on TOD$^3$Cap without any TOD$^3$Cap fine-tuning.
As shown in Table~\ref{tab:tod3cap_two_protocols} (bottom), 3D CoCa v2 achieves the best zero-shot performance and consistently improves over 3D CoCa.
These results indicate that the proposed inference-time optimization improves robustness under OOD shifts from indoor to outdoor scenes.
For completeness, Table~\ref{tab:tod3cap_two_protocols} (top) also reports the in-domain upper bound when methods are trained to convergence on TOD$^3$Cap.

\subsection{Ablation Study}
\begin{table}[!htbp]
    \centering
    \caption{
    \textbf{Effect of contrastive loss weight $\lambda$ on ScanRefer.} Standard decoding is used throughout (no TTS). The best CIDEr is achieved at $\lambda{=}1.0$.
    }
    \resizebox{0.8\linewidth}{!}{
    \begin{tabular}{ccccc}
        \toprule
        $\lambda$   & C@0.25$\uparrow$ & B-4@0.25$\uparrow$ & M@0.25$\uparrow$ & R@0.25$\uparrow$ \\
        \midrule
        0.0   & 74.12 & 40.98 & 27.45 & 58.76 \\
        0.1 & 77.30 & 41.80 & 28.10 & 59.60 \\
        0.5 & 79.55 & 42.55 & 28.75 & 60.40 \\ \midrule
        \textbf{1.0} & \textbf{85.42} & \textbf{45.56} & \textbf{30.95} & \textbf{61.98} \\ \midrule
        2.0 & 76.89 & 41.50 & 28.00 & 59.30 \\
        \bottomrule
    \end{tabular}
    }
    \label{tab:ablation_contr}
    \vspace{-0.4cm}
\end{table}

\noindent\textbf{Effect of the contrastive loss weight.}
We analyze the sensitivity of the backbone to the contrastive objective by varying the loss weight $\lambda \in \{0, 0.1, 0.5, 1.0, 2.0\}$ while using standard decoding throughout (no TTS).
As shown in Table~\ref{tab:ablation_contr}, removing the contrastive term ($\lambda{=}0$) leads to the lowest captioning scores.
Increasing $\lambda$ improves performance, with CIDEr@0.25 rising from 74.12 to 79.55 when $\lambda$ increases from 0 to 0.5.
The best overall results are obtained at $\lambda{=}1.0$.
Further increasing the weight to $\lambda{=}2.0$ degrades performance, suggesting that overly strong contrastive regularization can harm caption generation quality.

\noindent\textbf{Decoder architecture.}
We study the effect of the captioning decoder in the \emph{3D CoCa backbone} by replacing the proposed CoCa-style multimodal decoder with a GPT-2 captioner while keeping the scene encoder and training protocol unchanged.
All results are reported with standard decoding (no TTS).
As shown in Table~\ref{tab:ablation_decoder}, using the CoCa-style decoder yields consistently higher scores across all metrics.
This indicates that an explicitly cross-attentive multimodal decoder is beneficial for exploiting the contrastively aligned scene representations during caption generation.

\begin{table}[t]
    \centering
    \caption{
    \textbf{The impact of different caption generation decoders.} Comparison of the description indicators of the original GPT-2 generator and the CoCa-style multimodal decoder in this paper under the same visual features. Standard decoding is used throughout (no TTS).
    }
    \resizebox{1\linewidth}{!}{
    \begin{tabular}{lcccc}
        \toprule
        Decoder Setting & C@0.25$\uparrow$ & B-4@0.25$\uparrow$ & M@0.25$\uparrow$ & R@0.25$\uparrow$ \\
        \midrule
        GPT-2 Captioner (Baseline) & 76.20 & 41.00 & 27.80 & 59.50 \\ \midrule
        \textbf{CoCa Transformer (3D CoCa)} & \textbf{85.42} & \textbf{45.56} & \textbf{30.95} & \textbf{61.98} \\
        \bottomrule
    \end{tabular}
    }
    \label{tab:ablation_decoder}
    \vspace{-0.2cm}
\end{table}

\begin{table}[tbp]
    \centering
    \caption{
    \textbf{Comparison of the impact of different 3D point cloud encoder architectures on description performance. All results use standard decoding (no TTS).}
    }
    \resizebox{1\linewidth}{!}{
    \begin{tabular}{lcccc}
        \toprule
        Encoder Architecture & C@0.25$\uparrow$ & B-4@0.25$\uparrow$ & M@0.25$\uparrow$ & R@0.25$\uparrow$ \\
        \midrule
        PointNet++ (Baseline) & 72.48 & 38.95 & 26.80 & 56.30 \\ \midrule
        \textbf{Our scene encoder} & \textbf{85.42} & \textbf{45.56} & \textbf{30.95} & \textbf{61.98} \\
        \bottomrule
    \end{tabular}
    }
    \label{tab:ablation_encoder}
    \vspace{-0.4cm}
\end{table}
\noindent\textbf{3D scene encoder.}
We evaluate the contribution of the 3D scene encoder in the \emph{3D CoCa} by replacing the proposed point-tokenizer-based encoder (followed by a frozen CLIP vision transformer) with a conventional PointNet++ encoder~\cite{qi2017pointnet++}, while keeping the remaining components and training protocol unchanged.
All results are reported with standard decoding (no TTS).
As shown in Table~\ref{tab:ablation_encoder}, the proposed scene encoder achieves consistently higher captioning scores across all metrics, suggesting that the transformer-based tokenization and CLIP-informed representation provide a stronger interface for multimodal decoding.

\begin{figure*}[htbp]
    \centering
    \includegraphics[width=\linewidth]{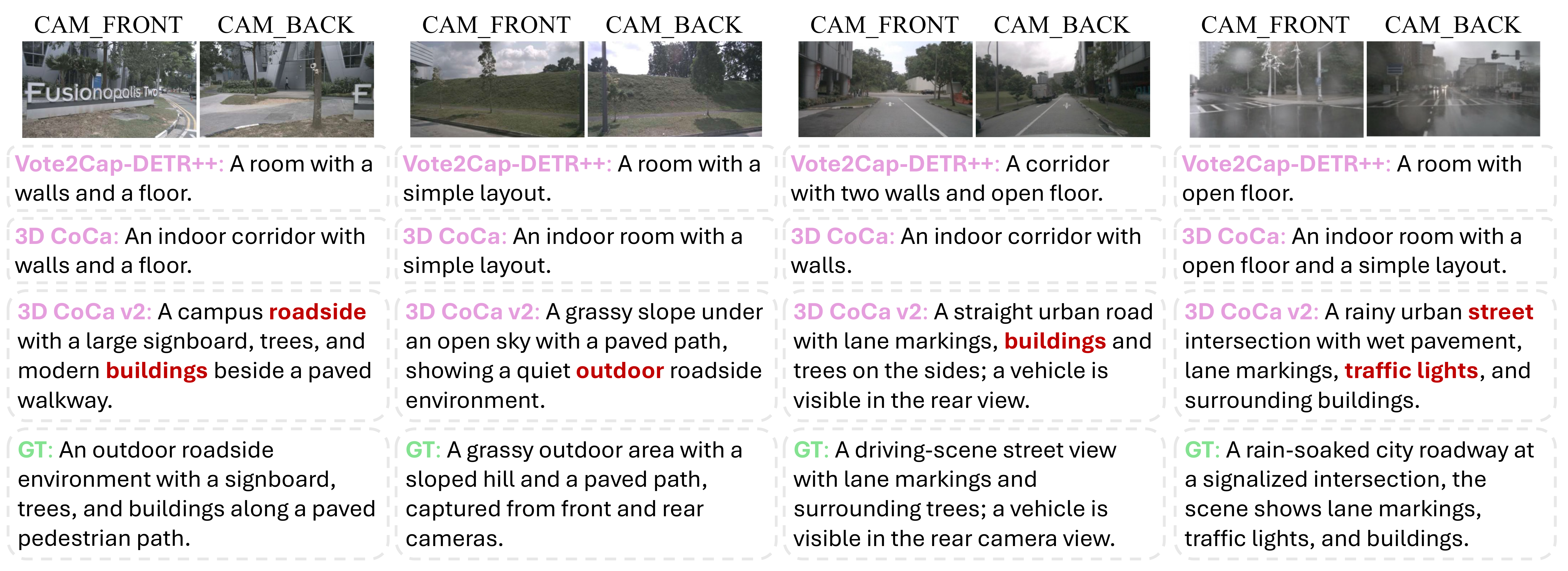}
    \caption{\textbf{Qualitative results on TOD$^3$Cap~\citep{todocap2025} (OOD, zero-shot).}
    We compare captions generated by the indoor-trained Vote2Cap-DETR++, 3D CoCa and 3D CoCa v2 on outdoor scenes with paired front and back views.
    Vote2Cap-DETR++ and 3D CoCa often exhibit a strong indoor bias, producing generic indoor descriptions, whereas 3D CoCa v2 generates more scene-consistent outdoor captions that better reflect key semantics. Ground-truth (GT) captions are shown for reference.
    Red words highlight informative details captured by 3D CoCa v2 but missing in the baseline.
    }
    \label{fig:visualization_ood}
    \vspace{-0.4cm}
\end{figure*}

\noindent\textbf{Ablation on the LLM judge.}
We study how the external judge $J(\cdot)$ affects TTS.
We fix the backbone, candidate generation (best-of-$N$, $N{=}8$), the compact scene summary $S(P)$, and the scoring prompt, and only vary the judge model.
As shown in Table~\ref{tab:ablation_judge_ref}, the relative ordering is consistent across datasets: stronger judges generally yield slightly higher CIDEr and lower hallucination rates.
In particular, GPT-5 achieves the best overall trade-off while also attaining the lowest hallucination rates.
Gemini~3 Pro is a close second, whereas lighter judges (Gemini~3 Flash and Qwen3-VL Pro) remain competitive with only a modest drop in CIDEr and a small increase in hallucinations.
These results indicate that TTS is not tied to a specific judge and can be paired with different LLMs to balance caption quality and faithfulness under varying inference budgets without updating the captioner parameters.
\begin{table}[t]
\centering
\caption{\textbf{Ablation on the LLM judge for TTS.}
We vary only the external judge $J$ while keeping the backbone, best-of-N decoding ($N{=}8$), scene summary $S(P)$, and the scoring prompt fixed. We report CIDEr@0.5$\uparrow$ and hallucination rate (Hall$\downarrow$, \%).
}
\label{tab:ablation_judge_ref}
\resizebox{\linewidth}{!}{
\begin{tabular}{l|cc|cc|cc}
\toprule
\multirow{2}{*}{Judge $J$ (TTS)} &
\multicolumn{2}{c|}{ScanRefer (in-domain)} &
\multicolumn{2}{c|}{Nr3D (in-domain)} &
\multicolumn{2}{c}{TOD$^3$Cap (OOD, zero-shot)} \\
\cmidrule{2-3} \cmidrule{4-5} \cmidrule{6-7}
& C@0.5$\uparrow$ & Hall$\downarrow$ & C@0.5$\uparrow$ & Hall$\downarrow$ & C@0.5$\uparrow$ & Hall$\downarrow$ \\
\midrule
Qwen3-VL Pro  & 78.00 & 9.2 & 53.70 & 9.8 & 49.20 & 13.6 \\
Qwen3-VL Max  & 78.30 & 8.9 & 54.00 & 9.5 & 49.80 & 12.9 \\
Gemini~3 Flash& 78.20 & 9.0 & 53.90 & 9.6 & 49.60 & 13.1 \\
Gemini~3 Pro  & 78.55 & 8.6 & 54.25 & 9.2 & 50.30 & 12.4 \\ \midrule
GPT-5 \textbf{(3D CoCa v2)}         & \textbf{78.63} & \textbf{8.2} & \textbf{54.45} & \textbf{8.8} & \textbf{50.80} & \textbf{11.8} \\
\bottomrule
\end{tabular}}
\vspace{-0.4cm}
\end{table}

\subsection{Qualitative Results}

\begin{figure*}[htbp]
    \centering
    \includegraphics[width=\linewidth]{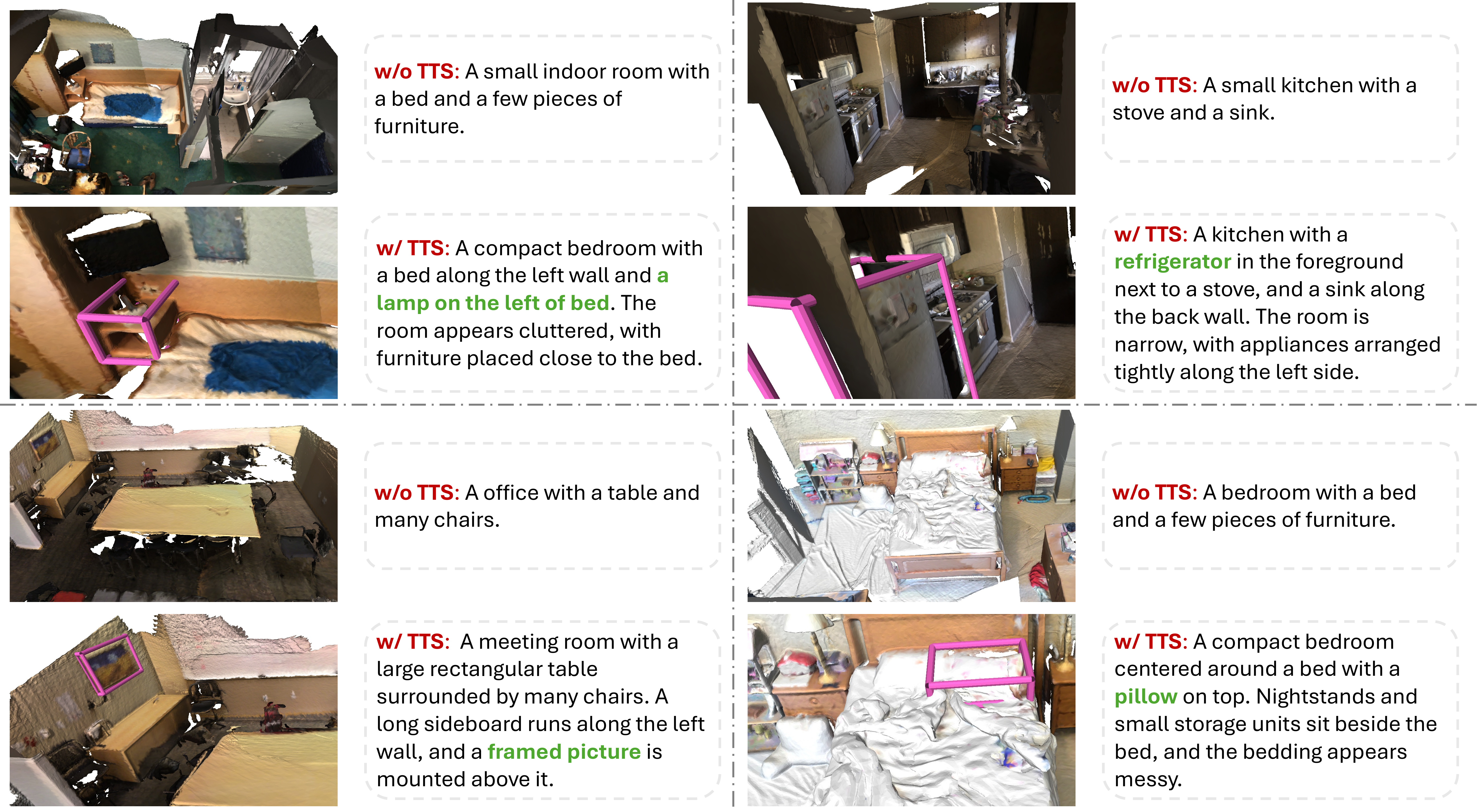}
    \caption{\textbf{Qualitative comparisons on ScanRefer~\cite{chen2020scanrefer} (w/o TTS vs w/ TTS).}
    For each example, we show the reconstructed 3D scene (top) and a zoomed-in view (bottom), where the target object is indicated by the \textcolor{magenta}{magenta} box.
    Compared with standard decoding (w/o TTS), Test-Time Search (w/ TTS) yields more specific and better-grounded captions, capturing object identities and layout cues supported by the highlighted region rather than generic room-level descriptions.
    \textcolor{green}{Green text} marks the object-specific details introduced by w/ TTS.
    }
    \label{fig:visual_comparison}
    \vspace{-0.4cm}
\end{figure*}

\noindent\textbf{In-domain qualitative results.}
Fig.~\ref{fig:visualization} presents qualitative comparisons on the \emph{in-domain} ScanRefer~\citep{chen2020scanrefer} benchmark.
Compared with 3D CoCa, 3D CoCa v2 generates captions that are more informative and better grounded, especially in cluttered indoor scenes.
As highlighted in red, our captions more frequently capture salient functional cues and fine-grained evidence (e.g., clutter attributes and object-level details), while avoiding generic or underspecified descriptions.

\noindent\textbf{Effect of test-time search (TTS).}
To further illustrate how TTS improves caption grounding, Fig.~\ref{fig:visual_comparison} shows side-by-side comparisons between standard decoding (w/o TTS) and our inference-only TTS (w/ TTS) on ScanRefer~\cite{chen2020scanrefer}.
The \textcolor{magenta}{magenta} box indicates the target object, and \textcolor{green}{green text} highlights object-specific details introduced by TTS.
Without TTS, the model tends to produce generic room-level descriptions.
In contrast, TTS favors candidates that are better supported by the highlighted region and the surrounding 3D context, resulting in more specific object identities and layout cues and fewer underspecified statements.

\noindent\textbf{OOD qualitative results.}
Fig.~\ref{fig:visualization_ood} further shows qualitative results on TOD$^3$Cap under OOD evaluation.
The indoor-trained Vote2Cap-DETR++ and 3D CoCa baseline exhibit a noticeable \emph{indoor bias}, often describing outdoor driving scenes using indoor concepts (e.g., ``room'' or ``corridor'') with generic layouts.
In contrast, 3D CoCa v2 produces captions that better match the outdoor context across front and back views, correctly emphasizing roads, buildings, trees, signboards, and vehicles.
Overall, these examples qualitatively support our OOD improvements, indicating that TTS mitigates domain-induced hallucinations and improves caption faithfulness under distribution shift.

\section{Test-Time Efficiency}
3D CoCa v2 adds an inference-time Test-Time Search that generates $N$ caption candidates and selects the best one using an external LLM judge.
Compared to standard decoding (TTS off; 3D CoCa), TTS incurs additional cost due to repeated decoding and judge scoring.
Table~\ref{tab:efficiency_latency} reports wall-clock latency per scene (batch size 1) with the default setting $N{=}8$.
TTS increases the total latency from 0.55s to 1.78s (3.24$\times$), while keeping the one-time backbone encoding cost unchanged (0.18s) and concentrating the overhead in the \emph{decode+judge} stage (1.60s).
Despite this overhead, the latency remains competitive with detector-heavy pipelines (e.g., 2.35 for Scan2Cap), and the added cost is justified by consistent gains in caption quality and faithfulness, particularly under OOD evaluation.
In practice, the quality-cost trade-off can be adjusted by $N$ and the choice of judge model.

\begin{table}[!htbp]
\centering
\caption{
\textbf{Test-time efficiency.}
Wall-clock latency per scene (s) with batch size 1.
\textit{Std.} denotes TTS disabled; \textit{TTS} uses best-of-$N$ with $N{=}8$ and the same summary and prompt as the main method.
}
\label{tab:efficiency_latency}
\resizebox{\linewidth}{!}{
\begin{tabular}{l|c|c|cc|c}
\toprule
\multirow{2}{*}{Method} & \multirow{2}{*}{Setting} & \multirow{2}{*}{Total$\downarrow$} &
\multicolumn{2}{c|}{Breakdown (s)$\downarrow$} & \multirow{2}{*}{Overhead$\downarrow$} \\
\cline{4-5}
 &  &  & Encode & Extra (dec+judge) &  \\
\midrule
3D CoCa~\citep{huang20253dcoca} & Std. (TTS off) & 0.55 & 0.18 & 0.37 & 1.00$\times$ \\
\textbf{3D CoCa v2 (Ours)} & \textbf{TTS ($N{=}8$)} & \textbf{1.78} & \textbf{0.18} & \textbf{1.60} & \textbf{3.24$\times$} \\
\midrule
Scan2Cap~\citep{scan2cap_2021} & Detector+Caption & 2.35 & 1.70 & 0.65 & 4.27$\times$ \\
Vote2Cap-DETR++~\citep{vote2cap++2024} & Detector+Caption & 2.80 & 2.10 & 0.70 & 5.09$\times$ \\
3D-VLP~\citep{3dvlp2024} & Retrieval-style & 2.05 & 1.55 & 0.50 & 3.73$\times$ \\
\bottomrule
\end{tabular}}
\vspace{-0.25cm}
\end{table}

\section{Limitation and Future Work}
3D CoCa v2 still has several limitations. TTS increases inference-time latency and may incur additional costs due to best-of-$N$ decoding and external judge queries, which can be undesirable for strict real-time deployment; although $N$ and the judge choice provide a controllable quality-efficiency trade-off, the overhead is not eliminated. Moreover, judge-guided selection depends on judge reliability and the fixed scoring prompt, and it can fail when the compact scene summary is incomplete or when the judge over-emphasizes fluency over grounding. Finally, our lightweight summary may miss fine-grained spatial relations or rare attributes, limiting its ability to penalize subtle hallucinations. Future work includes building more structured evidence representations, reducing TTS costs via adaptive $N$ and early stopping or learned lightweight judges, and extending the framework to broader 3D settings such as outdoor LiDAR, dynamic scenes, and embodied scenarios where captioning is coupled with actions and long-horizon memory.

\section{Conclusion}
We presented 3D CoCa v2, a unified contrastive-generative framework for 3D captioning that extends the 3D CoCa backbone with an inference-only TTS module.
TTS generates multiple diverse caption candidates and selects the best one via an external LLM judge conditioned on a compact scene summary, improving caption specificity and faithfulness without any additional training or parameter updates.
Extensive experiments on both in-domain indoor benchmarks and out-of-distribution outdoor scenes demonstrate that 3D CoCa v2 consistently enhances caption quality and robustness under distribution shift.
We hope this work highlights inference-time search as a practical, plug-and-play direction for building more reliable 3D captioners and motivates future research on stronger scene summarization, efficient judging, and generalizable 3D vision-language modeling for downstream embodied applications.

\noindent\textbf{Acknowledgements.}
This work was supported by the Fundamental Research Funds for the Central Universities, Peking University.

\bibliographystyle{plain}
\bibliography{template}   %

\end{document}